\documentclass[runningheads]{llncs}
\usepackage{graphicx}
\usepackage{multirow,makecell, rotating}
\usepackage{booktabs}
\usepackage{epsfig}
\usepackage{graphbox}
\usepackage{tikz}
\usepackage{comment}
\usepackage{amsmath,amssymb} 
\usepackage{color}
\usepackage[pagebackref,breaklinks,colorlinks]{hyperref}
\usepackage{booktabs}
\usepackage{bbm}
\usepackage{csquotes}

\usepackage[accsupp]{axessibility}  

\begin{document}
\pagestyle{headings}
\mainmatter
\def\ECCVSubNumber{263}  

\title{Anomaly Detection Requires Better Representations} 

\titlerunning{Anomaly Detection Requires Better Representations}
%
\author{Tal Reiss \and Niv Cohen \and Eliahu Horwitz \and Ron Abutbul \and Yedid Hoshen}
\authorrunning{T. Reiss et al.}
%
\institute{School of Computer Science and Engineering\\
  The Hebrew University of Jerusalem, Israel\\
\url{http://www.vision.huji.ac.il/ssrl_ad/} \\
}

\maketitle

\begin{abstract}
Anomaly detection seeks to identify unusual phenomena, a central task in science and industry. The task is inherently unsupervised as anomalies are unexpected and unknown during training. Recent advances in self-supervised representation learning have directly driven improvements in anomaly detection. In this position paper, we first explain how self-supervised representations can be easily used to achieve state-of-the-art performance in commonly reported anomaly detection benchmarks. We then argue that tackling the next generation of anomaly detection tasks requires new technical and conceptual improvements in representation learning.
\keywords{Anomaly Detection, Self-Supervised Learning, Representation Learning}
\end{abstract}

\section{Introduction}
\begin{quote}
\textit{Discovery commences with the awareness of anomaly, i.e., with the recognition that nature has somehow violated the paradigm-induced expectations that govern normal science.} 
\begin{flushright}
    \small{--------Kuhn, The Structure of Scientific Revolutions (1970)}
  \end{flushright}
\end{quote}

\begin{quote}
\textit{I do not know what I may appear to the world, but to myself I seem to have been only like a boy playing on the seashore, and diverting myself in now and then finding a smoother pebble or a prettier shell than ordinary, whilst the great ocean of truth lay all undiscovered before me.} 
\begin{flushright}
    \small{--------Isaac Newton}
  \end{flushright}
\end{quote}

Anomaly detection, discovering unusual patterns in data, is a core task for human and machine intelligence. The importance of the task stems from the centrality of discovering unique or unusual phenomena in science and industry. For example, the fields of particle physics and cosmology have, to large extent, been driven by the discovery of new fundamental particles and stellar objects. Similarly, the discovery of new, unknown, biological organisms and systems is a driving force behind biology. The task is also of significant economic potential. Anomaly detection methods are used to detect credit card fraud, faults on production lines, and unusual patterns in network communications.

Detecting anomalies is essentially unsupervised as only "normal" data, but no anomalies, are seen during training. While the field has been intensely researched for decades, the most successful recent approaches use a very simple two-stage paradigm: (i) each data point is transformed to a representation, often learned in a self-supervised manner. (ii) a density estimation model, often as simple as a $k$ nearest neighbor estimator, is fitted to the normal data provided in a training set. To classify a new sample as normal or anomalous, its estimated probability density is computed - low likelihood samples are denoted as anomalies. 

In this position paper, we first explain that advances in representation learning are the main explanatory factor for the performance of recent anomaly detection (AD) algorithms. We show that this paradigm essentially "solves" the most commonly reported image anomaly detection benchmark (Sec.~\ref{sec:representations}). While this is encouraging, we argue that existing self-supervised representations are unable to solve the next generation of AD tasks (Sec.~\ref{sec:bottlenecks}). In particular, we highlight the following issues: (i) masked-autoencoders are much worse for AD than earlier self-supervised representation learning (SSRL) methods (ii) current approaches perform poorly in datasets with multiple objects per-image, complex background, fine-grained anomalies. (iii) in some cases SSRL performs worse than handcrafted representations (iv) for "tabular" datasets, no representation performed better than the original representation of the data (i.e. that data itself) (v) in the presence of nuisance factors of variation, it is unclear whether SSRL can \textit{in-principle} identify the optimal representation for effective AD. 

Anomaly detection presents both rich rewards as well as significant challenges for representation learning. Overcoming these issues will require significant progress, both technical and conceptual. We expect that increasing the involvement of the self-supervised representation learning community in anomaly detection will mutually benefit both fields.

\section{Related Work}

Classical AD approaches were typically based on either density estimation \cite{eskin2002geometric,latecki2007outlier} or reconstruction \cite{pca}. 
With the advent of deep learning, classical methods were augmented by deep representations \cite{mathieu2015deep,zhang2016colorful,larsson2016learning,noroozi2016unsupervised}. A prevalent way to learn these representations was to use self-supervised methods, e.g. autoencoder \cite{ruff2018deep}, rotation classification \cite{golan2018deep,hendrycks2019using}, and contrastive methods \cite{csi,droc}. An alternative approach is to combine pretrained representations with anomaly scoring functions \cite{perera2019learning,mkd,panda,mean_shifted}. The best performing methods \cite{panda,mean_shifted} combine pretraining on auxiliary datasets and a second finetuning stage on the provided normal samples in the training set. It was recently established \cite{panda} that given sufficiently powerful representations (e.g. ImageNet classification), a simple criterion based on the $k$NN distance to the normal training data achieves strong performance. We therefore limit the discussion of AD in this paper to this simple technique. 

\section{Anomaly Detection as a Downstream Task for Representation Learning}
\label{sec:preliminary}

In this section we describe the computational task, method, and evaluation setting for anomaly detection.

\textbf{Task definition.} We assume access to $N$ random samples, denoted by ${\mathcal{X}}_{train} = \{x_1,x_2...x_N\}$, from the distribution of the normal data $p_{norm}(x)$. At test time, the algorithm observes a sample $\tilde{x}$ from the real-world distribution $p_{real}(x)$, which consists of a combination of the normal and anomalous data distributions: $p_{norm}(x)$ and $p_{anom}(x)$. The task is to classify the sample $\tilde{x}$ as normal or anomalous. 

\textbf{Representations for anomaly detection.} 
In AD, it is typically assumed that anomalies $a \sim p_{anom}$ have a low likelihood under the normal data distribution, i.e. that $p_{norm}(a)$ is small. Under this assumption, the PDF of normal data $p_{norm}$ acts as an effective anomaly classifier. In practice, however, training an estimator $q$ for scoring anomalies using $p_{norm}$ is a challenging statistical task. The challenge is greater when: (i) the data are high-dimensional (e.g. images) (ii) $p_{norm}$ is sparse or irregular (iii) normal and anomalous data are not separable using simple functions.
Representation learning may overcome these issues by transforming the sample $x$ into a representation $\phi(x)$, which is of lower dimension, where $p_{norm}$ is relatively smooth and where normal and anomalous data are more separable. As no anomaly labels are provided, self-supervised representation learning is needed.

\begin{figure*}[t]

\begin{center}
\includegraphics[width=1\linewidth]{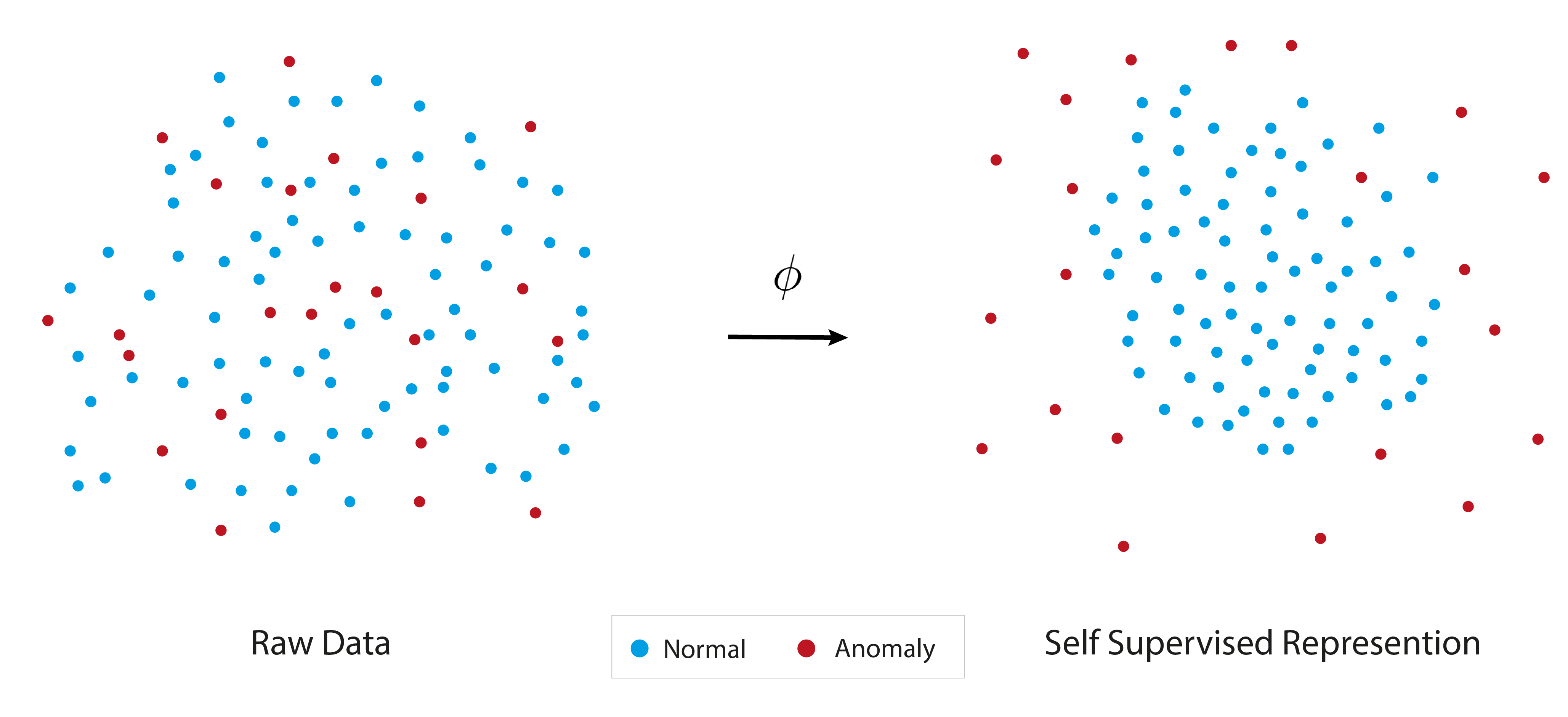}
\end{center}
\caption{\textit{\textbf{Normal and Anomalous Representations:}} The self-supervised representations transform the raw data into a space in which normal and anomalous data can be easily separated using density estimation methods}
\label{fig:normal_anomalous_rep}
\end{figure*}

\textbf{A two-stage anomaly detection paradigm.} Given a self-supervised representation $\phi$, we follow a simple two stage anomaly detection paradigm:
(i) \textit{Representation encoder}: each sample during training or test is mapped to a feature descriptor using the mapping function $\phi$.  
(ii) \textit{Density estimation}: a probability estimator $q_{norm}(x)$ is fitted to the distribution of the normal sample features ${\mathcal{X}}_{train} = \{\phi(x_1), \phi(x_2)...\phi(x_N)\}$. A sample is scored at test time by first mapping it to the representation space $\phi(\tilde{x})$ and scoring it according to the density estimator. Given an estimator $q_{norm}$ of the normal probability density $p_{norm}$, the anomaly score $s$ is given by $s(\tilde{x}) = -q_{norm}(\phi(\tilde{x}))$. Normal data will typically obtain lower scores than anomalous samples. A user can then set a threshold for the prediction of anomalies based on an appropriate false positive rate.

\section{Successful Representation Learning Enables Anomaly Detection}
\label{sec:representations}
Detecting anomalies in images is probably the most researched task by the deep learning anomaly detection community. In this section, we show that the simple paradigm presented in Sec.~\ref{sec:preliminary} achieves state-of-the-art results. As most density estimators achieve very similar results, the anomaly detection performance is mostly determined by the quality of learned representation. This makes anomaly detection an excellent testing ground for representations. Furthermore, we discuss different approaches to finetune a representation on the normal train data and show significant gains.

\textbf{Learning representations from the normal data.} Perhaps the most common approach taken by recent AD methods is to learn the representations in a self-supervised manner using solely the normal samples (i.e. the training dataset). Examples of such methods are RotNet \cite{hendrycks2019using}, CSI \cite{csi} and others. The main disadvantage of such methods is that most of the datasets are of small size and hence do not suffice for learning powerful representations. 

\textbf{Extracting representations from a pretrained model.} A very simple alternative is to use an off-the-shelf pretrained model and extract features for the normal (i.e. training) data from it. The pretraining may be either supervised (e.g. using ImageNet labels \cite{deng2009imagenet,he2016deep}) or self-supervised (e.g. DINO), in both cases pretraining may be performed on ImageNet. These representations tend to perform much better than those extracted from models trained only on the normal data.

\textbf{A hybrid approach.} A natural extension to the above approaches is to combine the two. This is done by using the pretrained model as an initialization for a self-supervised finetuning phase (on the normal data). In this way, the powerful representation of the pretrained model can be used and refined within the context of the anomaly detection dataset and task. Multiple approaches \cite{panda,mean_shifted} have been used for the self-supervised finetuning stage. However, in this paper we present what is possibly the simplest approach, using DINO's objective for the finetuning stage. In this approach, a pretrained DINO model is used as an initialization. During the finetuning phase, the model is trained on the target anomaly detection training dataset (i.e. only normal data) in a self-supervised manner by simply using the original DINO objective.

In Fig.~\ref{fig:normal_anomalous_rep} the above process is demonstrated with a toy example. Tab.~\ref{tab:dino} presents anomaly detection results on the CIFAR-10 \cite{krizhevsky2009learning} dataset, which is the most commonly used dataset for evaluation. As can be seen, using representations extracted from a recent self-supervised method (i.e. DINO) following the hybrid approach and coupled with a trivial $k$NN estimator for the density estimation phase nearly solves this dataset. Although a possible conclusion could have been that the anomaly detection task has been solved, in the next section we show this is not the case. 

\begin{figure}[t]
\centering
\begin{tabular}{@{\hskip2pt}c@{\hskip2pt}c@{\hskip2pt}|@{\hskip2pt}c@{\hskip2pt}c@{\hskip2pt}c@{\hskip2pt}c@{\hskip2pt}c}
& Image & \multicolumn{5}{c}{Nearest Neighbors}\\

\rotatebox[origin=c]{90}{MAE} &
\includegraphics[align=c, width=0.16\linewidth]{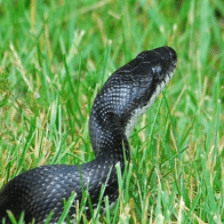} & 
\includegraphics[align=c, width=0.16\linewidth]{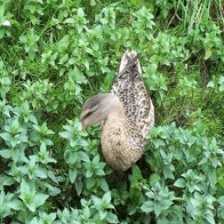} & 
\includegraphics[align=c, width=0.16\linewidth]{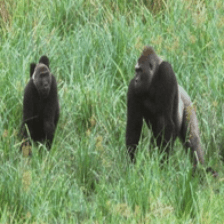} & 
\includegraphics[align=c, width=0.16\linewidth]{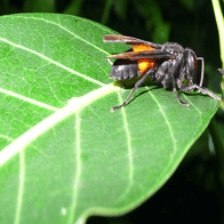} & 
\includegraphics[align=c, width=0.16\linewidth]{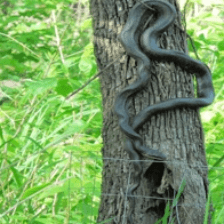} &
\includegraphics[align=c, width=0.16\linewidth]{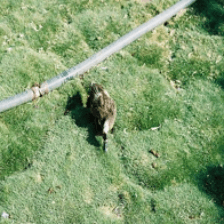} \\

\rotatebox[origin=c]{90}{DINO} &
\includegraphics[align=c, width=0.16\linewidth]{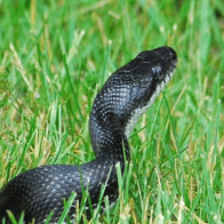} & 
\includegraphics[align=c, width=0.16\linewidth]{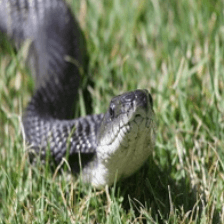} & 
\includegraphics[align=c, width=0.16\linewidth]{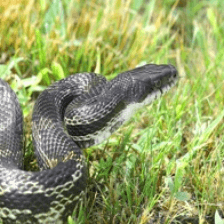} & 
\includegraphics[align=c, width=0.16\linewidth]{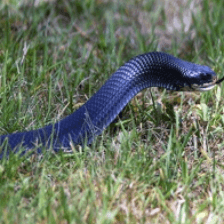} & 
\includegraphics[align=c, width=0.16\linewidth]{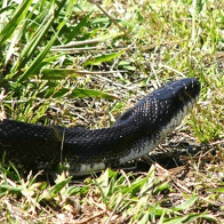} &
\includegraphics[align=c, width=0.16\linewidth]{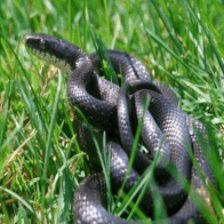} \\
\\
\rotatebox[origin=c]{90}{MAE} &
\includegraphics[align=c, width=0.16\linewidth]{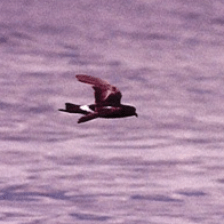} & 
\includegraphics[align=c, width=0.16\linewidth]{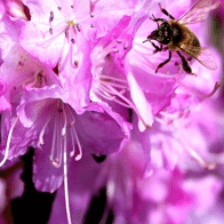} & 
\includegraphics[align=c, width=0.16\linewidth]{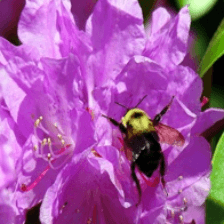} & 
\includegraphics[align=c, width=0.16\linewidth]{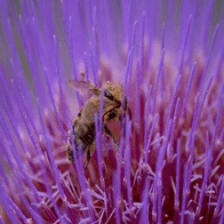} & 
\includegraphics[align=c, width=0.16\linewidth]{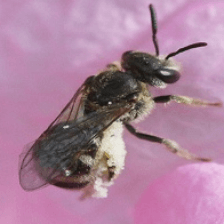} &
\includegraphics[align=c, width=0.16\linewidth]{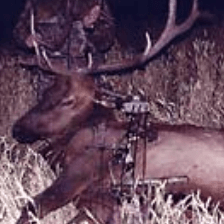} \\ 

\rotatebox[origin=c]{90}{DINO} &
\includegraphics[align=c, width=0.16\linewidth]{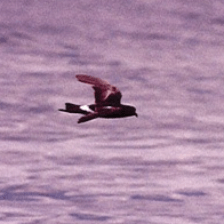} & 
\includegraphics[align=c, width=0.16\linewidth]{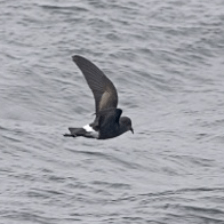} & 
\includegraphics[align=c, width=0.16\linewidth]{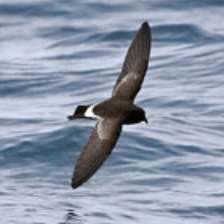} & 
\includegraphics[align=c, width=0.16\linewidth]{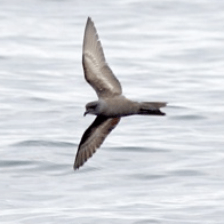} & 
\includegraphics[align=c, width=0.16\linewidth]{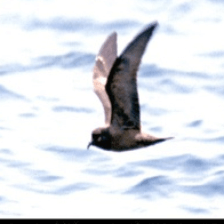} &
\includegraphics[align=c, width=0.16\linewidth]{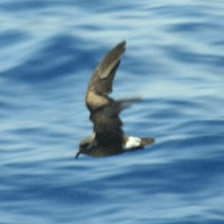} \\ 

\end{tabular}
\caption{\textit{\textbf{MAE vs. DINO nearest neighbors:}} For each image, the top $5$ nearest neighbors are shown according to their order. Note how MAE neighbors are chosen mostly based on the colors and not their semantic contents, in contrast, DINO neighbors are semantically accurate.}
\label{fig:nn_images}
\end{figure}

\begin{table}[t]
  \caption{\textit{\textbf{Image anomaly detection results:}} Mean ROC-AUC \%. Bold denotes the best results, FT stands for finetuned}
  \label{tab:dino}
  \centering
  \begin{tabular}{lccccccccccccccc}
    \toprule
Approach	&	\multicolumn{2}{c}{Self-supervised}  & \multicolumn{2}{c}{Pretrained} & \multicolumn{3}{c}{Hybrid} \\
    \cmidrule(r){1-1}
    \cmidrule(r){2-3}
    \cmidrule(r){4-5}
    \cmidrule(r){6-8}
Method	& \multicolumn{1}{c}{RotNet \cite{hendrycks2019using}} & CSI \cite{csi}& \multicolumn{1}{c}{ResNet} & \multicolumn{1}{c}{DINO} & \multicolumn{1}{c}{PANDA \cite{panda}}  & \multicolumn{1}{c}{MSAD} \cite{mean_shifted} &  \multicolumn{1}{c}{DINO-FT} \\

    \cmidrule(r){1-1}
    \cmidrule(r){2-2}
    \cmidrule(r){3-3}
    \cmidrule(r){4-4}
    \cmidrule(r){5-5}
    \cmidrule(r){6-6}
    \cmidrule(r){7-7}
    \cmidrule(r){8-8}

CIFAR-10  &  90.1 & 94.3 & 92.5 & 97.1 & 96.2 & 97.2 & \textbf{98.4} \\	

    \bottomrule
  \end{tabular}
\end{table}

\section{Gaps in Anomaly Detection Point to Bottlenecks in Representations Learning}
\label{sec:bottlenecks}

While Sec.~\ref{sec:representations} presented a very optimistic view of the ability of representation learning to solve anomaly detection, in this section we paint a more complex picture. We use this to highlight several limitations of current self-supervised representations. 

\subsection{Masked-Autoencoder: Advances in self-supervised learning do not always imply better anomaly detection}
\label{subsec:mae}
Recently, masked-autoencoder (MAE) based methods achieved significant improvements on several self-supervised representation learning benchmarks \cite{mae}. Yet, the representations learnt by MAE underperform contrastive self-supervised methods on unsupervised tasks such as anomaly detection. A comparison between MAE to contrastive self-supervised method (DINO) is presented in Tab.~\ref{tab:dino-mae} demonstrating the much better performance of DINO for AD. Finetuning on the normal training data improves both methods, however a large gap still remains. Implementation details for the experiments can be found in the App.~\ref{sec:appendix}.  In many papers, self-supervised methods are evaluated using supervised benchmarks, such as classification accuracy with finetuning. The key difference between anomaly detection and ordinary benchmarks where MAE excel is that anomaly detection is an unsupervised task. This is also suggested by MAE's worse performance with linear probing (as reported by the original paper), where the supervised labels cannot be used to improve the backbone representations. 

MAE's optimization objective may explain why its strong representation does not translate into better anomaly detection capabilities. As MAE's objective is to reconstruct patches, it may learn a representation that encodes local information needed for reconstructing the image, overlooking semantic object properties. Consequently, the nearest neighbors may pay more attention to local similarity than to global semantic properties (See Fig.~\ref{fig:nn_images}). In contrast, the goal of contrastive-based objectives is to map semantically similar images to nearby representations, disregarding some of the local properties.


\textbf{Conclusion.} Better performance on supervised downstream tasks does not necessarily imply  better representations across the board. In some cases, while the representation may excel in a supervised downstream task, it may underperform in an unsupervised counterpart. Looking forward, we suggest that new self-supervised representation learning methods present evaluations on unsupervised anomaly detection tasks alongside the common supervised benchmarks. 

\begin{table}[ht]
\caption{\textit{\textbf{Anomaly detection comparison of MAE and DINO:}} Mean ROC-AUC \%. Bold denotes the best results}
\centering
\label{tab:dino-mae}
\begin{tabular}{lccc}
\toprule 
Method & CIFAR-10 & CUB-200 & INet-S \\ 
    \cmidrule(r){1-1}
    \cmidrule(r){2-2}
    \cmidrule(r){3-3}
    \cmidrule(r){4-4}
MAE       &  78.1  & 73.1 & 83.2 \\
DINO     &  \textbf{97.1}  & \textbf{93.9} & \textbf{99.3}  \\
\bottomrule
\end{tabular}

\end{table}

\subsection{Complex datasets: Current representations struggle on scenes, finegrained classes, multiple objects}

Current representations are very effective for anomaly detection on datasets with a single object occupying a large portion of the image. Furthermore, these methods typically perform well when the number of object categories in the normal train set is small and have coarse differences (e.g. "cat" and "ship"). A prime example is CIFAR-10, which is virtually solved. On the other hand, anomaly detection accuracy is much lower on more complex datasets containing multiple small objects, complex backgrounds; and when anomalies consist of related object categories (e.g. "sofa" and "armchair"). We modified the MS-COCO \cite{lin2014microsoft} dataset by using all images from a single super-category ('vehicles') as normal data, apart from a single category ('bicycle') which are used as anomalies. We experiment both with cropping just the object bounding boxes or using the entire image (including the background and other objects). Similarly, we report results for a multi-modal CUB-200 \cite{welinder2010caltech} anomaly detection benchmark. The results are presented in Tab.~\ref{tab:ms-coco} (implementation details can be found in the Appendix). It is clear that these datasets are far from solved and that better representations are needed to achieve acceptable performance.

\textbf{Conclusion.} While current representations are effective for relatively easy datasets, more realistic cases with small objects, backgrounds and many object categories call for the development of new SSRL methods.

\begin{table}
\caption{\textit{\textbf{Multi-modal datasets:}} Mean ROC-AUC \%. ”MS-COCO-I” / ”MS-COCO-O” indicates MS-COCO image / object level benchmarks (respectively).}
\centering
\begin{tabular}{lccc}
\toprule
Method	&	MS-COCO-I	&	MS-COCO-O	&	CUB-200		\\ 		
\cmidrule(r){1-1}\cmidrule(r){2-2}\cmidrule(r){3-3}\cmidrule(r){4-4}
PANDA \cite{panda}	&	61.5	&	77.0	&   78.4    \\
MSAD \cite{mean_shifted}	&	61.7	&	76.9	&   80.1	\\
DINO    &   61.5    &   73.4    &   74.5    \\
\bottomrule
\end{tabular}
\label{tab:ms-coco}
\end{table}

\subsection{Unidentifiability: Representations for anomaly detection may be ambiguous without further guidance}

In some settings, we would like our representation to focus on specific attributes (which we denote as \textit{relevant}) while ignoring nuisance attributes that might bias the model. Consider two different companies interested in anomaly detection in cars. The first company may be interested in detecting novel car models, while the second is interested in unusual driving behaviors. Although both may wish to apply density estimation using a state-of-the-art self-supervised representation, each will view the ground truth anomalies of the other company as a false-positive case. As each company is interested in different anomalies, they may require different representations. One company would require the representation to contain only the driving patterns and be agnostic to the car model, at the same time, the other company would strive for the opposite. As these preferences are not present at the time of pretraining the self-supervised backbone, the correct solution is often unidentifiable.

One initial effort is RedPANDA \cite{cohen2022red} that proposed providing labels for nuisance attributes. We note that only the attributes to be ignored are labeled, while the other attributes (the ones in which characterize anomalies) are not provided. Representation learning is then performed using domain-supervised disentanglement \cite{kahana2022contrastive}, resulting in a representation only describing the unlabelled attributes. Yet, the field of domain-supervised disentanglement is still in its infancy, and the assumption of nuisance attribute labels is often not applicable. 

\textbf{Conclusion.}
Self-supervised representation learning methods are designed to focus on semantic attributes of images, but choosing the most relevant ones is unidentifiable without further guidance. Incorporating guidance may be achieved by a careful choice of inductive bias \cite{kahana2022contrastive} (e.g. augmentations) or using concept-based representation techniques \cite{koh2020concept}.

\begin{table}[t]
\begin{center}
\caption{\textit{\textbf{Summary of the findings of from Horwitz and Hoshen}}~\cite{3d_ads}: Average metrics across all MVTec3D-AD classes, "INet" indicates ImageNet \cite{deng2009imagenet} pretrained features, PC indicates point cloud. I-ROC  indicates image level ROC-AUC \% \cite{rocauc}, P-ROC indicates pixel level ROC-AUC \%. Higher score indicates better the results}
\label{table:3d_rep_pro}

\begin{tabular}{l@{\hskip5pt}|c@{\hskip5pt}c@{\hskip5pt}c@{\hskip5pt}c@{\hskip5pt}c@{\hskip5pt}c@{\hskip5pt}c@{\hskip5pt}c@{\hskip5pt}} 
    \toprule
    Modality & RGB & Depth & Depth & Depth & Depth & Depth  & PC & RGB+PC \\
    Method & INet & INet & NSA \cite{nsa} & Raw & HoG \cite{hog} & SIFT \cite{sift} & FPFH \cite{fpfh} & RGB+FPFH\\
    \midrule
    PRO \cite{mvtec2d} & 87.6 & 58.6 & 57.2 & 19.1 & 61.4 & 86.6 & 92.4 &\textbf{96.4}\\
    I-ROC  & 78.5 & 63.7 & 69.6 & 52.8 & 56.0 & 71.4 & 75.3 & \textbf{86.5}\\
    P-ROC  & 96.6 & 82.1 & 81.7 & 54.8 & 84.5 & 95.4 & 98.0 &\textbf{99.3}\\
   \bottomrule
   
    \end{tabular}
    
\end{center}
\end{table}

\subsection{3D Point Clouds: Self-supervised representations do not always improve over handcrafted ones}
In an empirical investigation \cite{3d_ads}, we evaluated representative methods designed for different modalities on the MVTec3D-AD dataset \cite{mvtec3d_ad}. The paper showed that currently, handcrafted features for 3D surface matching outperform learning-based methods designed either for images or for 3D point clouds. A key insight was that rotation invariance is very beneficial in this modality, and is often overlooked. A summary of the findings, taken from the original paper, is found in Tab.~\ref{table:3d_rep_pro}.

\textbf{Conclusion.}
When dealing with 3D point-cloud, self-supervised representations are yet to outperform handcrafted features for anomaly detection. For modalities less mature than images, domain specific priors may still need to be integrated into the architecture or objective. This stresses the need for better 3D point-cloud representations.

\subsection{Tabular Data: When representations do not improve over the original data}
\label{subsec:tabular_domain}
The tabular setting is probably the most general anomaly detection setting, where each sample in the dataset consists of a set of numerical and categorical variables. This is strictly harder than any other setting as no regularity in the data can be assumed. Such data are frequently encountered, as unstructured databases are very common. In recent years, self-supervised methods have been proposed for tabular anomaly detection~\cite{zong2018deep,bergman2020classification,icl,qiu2021neural}. These methods differ by the auxiliary task that they use for representation learning (and potentially also for anomaly scoring). Two representative deep learning approaches are GOAD~\cite{bergman2020classification} which predicts geometric transformations, and use the prediction errors to detect anomalies, and ICL~\cite{icl} which adopts the contrastive learning task for training and for anomaly scoring by differentiating between in-window and out-window features. As part of our evaluation, we used both their standard pipeline (i.e. their auxiliary tasks for anomaly scoring) and our AD density estimation paradigm (see Appendix). These results were then compared with $k$NN on the original raw features without any modifications. The results are presented in Tab.~\ref{tab:tabular}. Self-supervised representation learning did not improve performance in comparison with the original raw features. 


\textbf{Conclusion.}
Representation learning for general datasets is an open research question. Some prior knowledge of the dataset must be used in order to learn non-trivial data representations, at least in the context of anomaly detection. 

\begin{table}[t]
  \caption{\textit{\textbf{Tabular results:}} Mean F1 \& ROC-AUC \% from the ODDS benchmarks results. Bold denotes the best results}
  \label{tab:tabular}
  \centering
  \begin{tabular}{lcccccccccccccccccccc}
    \toprule
Method	&	\multicolumn{2}{c}{GOAD \cite{bergman2020classification}}  & \multicolumn{2}{c}{ICL \cite{icl}} & \multicolumn{1}{c}{Raw}  \\

    \cmidrule(r){1-1}
    \cmidrule(r){2-3}
    \cmidrule(r){4-5}
    \cmidrule(r){6-6}
Scoring & \multicolumn{1}{c}{Auxiliary} & \multicolumn{1}{c}{\textit{k}NN} & \multicolumn{1}{c}{Auxiliary} & \multicolumn{1}{c}{\textit{k}NN} & \multicolumn{1}{c}{\textit{k}NN}\\
 
     \cmidrule(r){1-1}
    \cmidrule(r){2-2}
    \cmidrule(r){3-3}
    \cmidrule(r){4-4}
    \cmidrule(r){5-5}
    \cmidrule(r){6-6}

F1	&  54.4	& 63.2 & 68.1 & 69.8 & \textbf{69.9}\\	
ROC-AUC	&  78.2 & 87.6 & 88.9 & 89.4 & \textbf{90.2}\\	

    \bottomrule
  \end{tabular}
\end{table}


\section{Final Remarks}
In this position paper, we advocated the study of self-supervised representations for the task of anomaly detection. We explained that advances in representation learning have been the main driving force behind progress in anomaly detection. On the other hand, we demonstrated that current self-supervised representation learning methods often fall short in challenging anomaly detection settings. Our hope is that interplay between the self-supervised representation learning and anomaly detection fields will result in mutual benefits for both communities.

\section{Acknowledgements}
This work was partially supported by the Malvina and Solomon Pollack Scholarship, a Facebook award, the Israeli Cyber Directorate, the Israeli Higher Council and the Israeli Science Foundation. We also acknowledge support of Oracle Cloud credits and related resources provided by the Oracle for Research program.

\appendix

\section{Appendix}
\label{sec:appendix}
In this paper we report anomaly detection results using the standard uni-modal protocol, which is widely used in the anomaly detection community. In the uni-modal protocol, multi-class datasets are converted to anomaly detection by setting a class as normal and all other classes as anomalies. The process is repeated for all classes, converting a dataset with $C$ classes into $C$ datasets. Finally, we report the mean ROC-AUC \% over all $C$ datasets as the anomaly detection results.

\subsection{Anomaly detection comparison of MAE and DINO} 
We compare between DINO \cite{dino} and MAE \cite{mae} as a representation for a $k$NN based anomaly detection algorithm. For MAE, we experimented both with $k$NN and reconstruction error for anomaly scoring and found that the latter works badly, therefore we report just the $k$NN results. We evaluate using a variety of datasets, in the uni-modal setting described above. We used the following datasets:

\textit{INet-S} \cite{ridnik2021imagenet21k}: The dataset is subset of 10 animal classes taken from ImageNet21k (e.g "petrel", "tyrannosaur", "rat snake", "duck", "bee fly", "sheep", "beer cub", "red deer", "silverback", "opossum rat") that do not appear in ImageNet1K dataset. The dataset is coarse-grained and contains images relatively close to ImageNet1K dataset. It intended to convey that even for easy tasks the MAE doesn't achieve as good results as DINO.

\textit{CIFAR-10} \cite{krizhevsky2009learning}:
Consists of low-resolution $32\times32$ images from 10 different classes.

\textit{CUB-200} \cite{welinder2010caltech}:
Bird species image dataset which contains 11,788 images of 200 subcategories. In the experiment we calculated mean ROC-AUC \% over the 20 first categories.

\subsection{Multi-modal datasets} 
In these experiment we specify a single class as  anomalous, and treat all images which does not contain it as normal.

\textit{MS-COCO-I} \cite{lin2014microsoft}: We build a multi-modal anomaly detection dataset comprised of scenes benchmarks, where each image is evaluated against other images featuring similar scenes.
We choose 10 object categories ("bicycle","traffic light", "bird" , "backpack", "frisbee", "bottle",  "banana",  "chair", "tv",  "microwave", "book") from different MS-COCO super-categories. To construct a multi-modal anomaly detection benchmark, we designate an object category from the list as the anomalous class, and training images of a similar super-category that do not contain it as our normal train set. Our test set contains all the test images from that super-category, where images containing the anomalous object are labelled as anomalies. This process is repeated for the 10 object categories resulting in 10 different evaluations. We report their average ROC-AUC \%.

\textit{MS-COCO-O}: We introduce a similar benchmark to MS-COCO-I, focusing on single objects rather than scenes. We crop all objects from our 10 super-categories (described above) according to the MS-COCO supplied bounding boxes. We repeat a similar process, using a similar object category as normal and the rest as anomalies.

\textit{CUB-200} \cite{welinder2010caltech}:
We create a multi-modal anomaly detection benchmark based on the CUB-200 dataset. We focus on the 20 first categories, designating only one as an anomaly each time.

\subsection{Tabular domain}
Various datasets used for tabular data anomaly detection were used for the experiments. A total of 31 datasets from Outlier Detection DataSets (ODDS)\footnote{http://odds.cs.stonybrook.edu/} are employed. 
For the evaluation of GOAD and ICL we used the official repositories and made an effort to select the best configuration available. For all density estimation evaluations we used $k$NN with $k=5$ nearest neighbors. To convert GOAD and ICL into the standard paradigm of representation learning followed by density estimation: i) we use the original approaches to train a feature encoder (followed by a classifier which we discard) ii) we use the feature encoder to represent each sample iii) density estimation is performed on the representations using $k$NN exactly as in Sec.~\ref{sec:preliminary}.

\clearpage
%
%
\bibliographystyle{splncs04}
\bibliography{egbib}
\end{document}